\documentclass{ICRES}

\begin{document}

\title{RoboTed: a case study in Ethical Risk Assessment}

\author{Alan F.T. Winfield and Katie Winkle}

\address{Bristol Robotics Laboratory, UWE Bristol, UK\\
E-mail: alan.winfield/katie.winkle@brl.ac.uk\\
www.brl.ac.uk}

\begin{abstract}
Risk Assessment is a well known and powerful method for discovering and mitigating risks, and hence improving safety. Ethical Risk Assessment uses the same approach but extends the envelope of risk to cover ethical risks in addition to safety risks. In this paper we outline Ethical Risk Assessment (ERA) and set ERA within the broader framework of Responsible Robotics. We then illustrate ERA with a case study of a hypothetical smart robot teddy bear: RoboTed. The case study shows the value of ERA and how consideration of ethical risks can prompt design changes, resulting in a more ethical and sustainable robot.
\end{abstract}


\bodymatter

\section{Introduction}

Risk Assessment is a well known method for discovering and mitigating risks, and hence improving safety. Ethical Risk Assessment is not new either; it is essentially what research ethics committees do\cite{bern2012}. But the idea of extending the envelope of safety risk assessment of intelligent systems to encompass ethical risks is new. Given the growing awareness of the ethical risks of intelligent systems in recent years, ethical risk assessment offers a powerful method for systematically identifying and mitigating the ethical, societal and environmental risks associated with the use of robots and artificial intelligence (AI).

In this paper we first define ethical risk assessment (ERA) with reference to British Standard BS8611 then, in section 3, present a worked example of ERA based upon a hypothetical smart robot teddy bear we call RoboTed. The paper concludes with an appraisal of both the benefits and limitations of ERA.

\section{Ethical Risk Assessment}

Risk Assessment is a process which typically has three stages:
\begin{enumerate}
    \item identify and analyse potential events (hazards) that may cause harm to individuals, property, and/or the environment;
    \item make judgments on the acceptability and likely impact of the harm arising from exposure to the hazard (risks), then
    \item determine what steps should be taken to mitigate those risks and hence minimise or eliminate possible harms.
\end{enumerate}

Standards for risk assessment are well established in safety critical systems. ISO 14971:2007 \textit{Application of risk management to medical devices}, for instance, provides requirements and guidance on risk assessment for medical devices.  And ISO 12100:2010 \textit{Safety of machinery -- Risk assessment and risk reduction} sets out requirements on performing risk assessments, notably including risk analysis focused on hazard identification.

Almost certainly the world's first explicitly ethical standard in robotics is BS8611-2016 \textit{Guide to the ethical design and application of robots and robotic systems}. ``BS8611 is not a code of practice, but instead guidance on how designers can undertake an ethical risk assessment of their robot or system, and mitigate any ethical risks so identified. At its heart is a set of 20 distinct ethical hazards and risks, grouped under four categories: societal, application, commercial and financial, and environmental. Advice on measures to mitigate the impact of each risk is given, along with suggestions on how such measures might be verified or validated" \cite{winf19}. Societal hazards include, for example, anthropomorphisation, loss of trust, deception, infringements of privacy \& confidentiality, addiction, and loss of employment.

BS8611 defines an \textbf{ethical harm} as ``anything likely to compromise psychological and/or societal and environmental well-being". An \textbf{ethical hazard} as ``a potential source of ethical harm" and an \textbf{ethical risk} as the ``probability of ethical harm occurring from the frequency and severity of exposure to a hazard" \cite{bsi16}. Ethical risk assessment thus extends the envelope of risk assessment to include ethical harms, hazards and risks (in addition to physical harms, hazards and risks).

Ethical Risk Assessment is part of the practice of Responsible Robots, which we define as ``the application of Responsible Innovation\cite{STILGOE20131568} in the design, manufacture, operation, repair and end-of-life recycling of robots, that seeks the most benefit to society and the least harm to the environment" \cite{winf2020}. We would expect ERA to be undertaken within a framework of responsible innovation -- such as EPSRC's AREA framework\footnote{\url{https://epsrc.ukri.org/research/framework/area/}} -- and alongside ethically aligned\cite{ieee19} and values-based design\cite{spiek20}.

\section{RoboTed}

\begin{figure}[htbp]
\centering
\includegraphics[width=6.0cm]{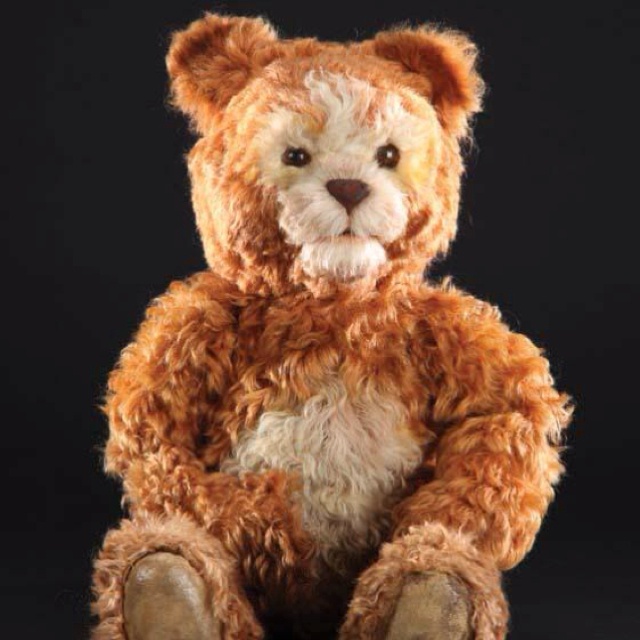}
\caption{Teddy}
\label{Fig:Teddy}
\end{figure}

Our hypothetical robot teddy bear is inspired by the robot \textit{Teddy} in the 2001 movie \textit{A.I. Artificial Intelligence}, directed by Steven Spielberg\footnote{as a tribute to Stanley Kubrick.}. \textit{Teddy}, shown in Fig.\ref{Fig:Teddy}, is far more advanced in intelligence, mobility and longevity than we can presently contemplate, so let us instead specify a robot teddy bear closer to today's capabilities.

Our fictional RoboTed is designed to:
\begin{enumerate}
\item Recognise its owner, learning their face and name, turning its face toward the child,
\item respond to physical play such as hugs and tickles,
\item tell stories, while allowing a child to interrupt the story to ask questions or ask for sections to be repeated,
\item sing songs, while encouraging the child to sing along and learn the song, and
\item act as a child minder, allowing parents to remotely listen, watch and speak via RoboTed.
\end{enumerate}

And is based upon the following technology:
\begin{enumerate}
\item RoboTed is an Internet (WiFi) connected device,
\item RoboTed has cloud-based speech recognition and conversational AI (chatbot) and local speech synthesis,
\item RoboTed’s eyes are functional cameras allowing the robot to recognise faces, and
\item RoboTed has motorised arms and legs to provide it with limited baby-like movement and locomotion -- not walking but shuffling and crawling. 
\end{enumerate}

\subsection{Assessing the Ethical Risks of RoboTed}

As a worked example we now consider the ethical hazards and risks of RoboTed, under the four categories of: physical (safety) risks, psychological risks, privacy \& security risks, and environmental risks.


\subsubsection{Physical Risks}

We first consider physical hazards and risks, summarized here in Table \ref{tab:1}, together with an assessment of the likelihood of occurrence of the risk (High, Medium or Low).

\begin{table}
\tbl{Physical Risk Analysis for RoboTed}
{\begin{tabular}{p{2cm} p{4cm} p{1cm} p{4cm}}\toprule
Hazard & Risk & Level$^{\text a}$ & Mitigation  \\
\colrule
Tripping & User(s) trip over RoboTed when it is crawling on the floor & M & Audible crawling sound to alert users to its presence\\
Battery overheating & Defective batteries or battery chargers can overheat or \textit{in extremis} catch fire & M & Design to make use of consumer rechargeable batteries rather than high-risk Li-Ion batteries.\\
\botrule
\end{tabular}
}
\begin{tabnote}
$^{\text a}$Risk level: (H)igh, (M)edium or (L)ow\\
\end{tabnote}
\label{tab:1}
\end{table}

\paragraph{Tripping}
As RoboTed crawls on the floor, it has the potential to become a trip hazard. A mitigation strategy might be to have RoboTed make an audible crawling sound when it is moving, to alert users (particularly adults) to its presence.

\paragraph{Battery overheating}
There is a risk that defective batteries or battery chargers can overheat or \textit{in extremis} catch fire. In mitigation RoboTed should be designed to make use of low-risk consumer rechargeable batteries rather than high-risk Li-Ion batteries. In addition parents should be advised to supervise battery re-charging.

\subsubsection{Psychological Risks}

We now consider the psychological hazards and risks summarised in Table \ref{tab:2}.

\begin{table}
\tbl{Psychological Risk Analysis for RoboTed}
{\begin{tabular}{p{2cm} p{4cm} p{1cm} p{4cm}}\toprule
Hazard & Risk & Level$^{\text a}$ & Mitigation  \\
\colrule
Addiction & Child plays with RoboTed obsessively and neglects family & M & Explore `RoboTed needs to sleep now' function\\
Deception (of child) & Child believes that RoboTed has feelings (for her) & M & Design chatbot to avoid language that suggests feelings \\
Over trusting (by child) & Child cannot distinguish mode of operation & H & Notification when child minder mode activated, uses parents' voice rather than RoboTed's voice \\
Over trusting (by parents) & Parents come to rely on the child minder function & H & Remove the child minder function\\
The Uncanny Valley & Child becomes fearful of robot & L & Use ‘cartoon’ voice; engage children in early user trials\\
\botrule
\end{tabular}
}
\begin{tabnote}
$^{\text a}$(H)igh, (M)edium or (L)ow\\
\end{tabnote}
\label{tab:2}
\end{table}

\paragraph{Addiction} 
RoboTed might be so compelling that it leads to a child playing obsessively with RoboTed and neglecting his or her family \cite{sharkey2010crying}. This also increases the risk of emotional distress should RoboTed's behaviour change or fail in any way (e.g. if the facial recognition was to fail and no longer recognises the child). A mitigation strategy might be to explore the addition of a ‘RoboTed needs to sleep’ function, as a way of limiting length of play times.

\paragraph{Deception} 
There is a risk that the child comes to believe that RoboTed has feelings for her \cite{sparrow2006hands}. To mitigate this risk we could design the chatbot to avoid language that suggest feelings, so that RoboTed never says things like “I like you” or “why are you sad?”.

\paragraph{Over trusting by the child}
Building on deception there is a risk that the child cannot tell whether RoboTed is operating autonomously or is in child minder mode. This may result in her sharing sensitive information in the belief that no one else will hear it, when actually her parents are watching and listening. The reverse is also true, in that she may share something she wishes her parents to know but is too embarrassed to raise face to face, when actually the robot is operating autonomously and her parents are \textit{not} listening. Mitigation strategies would be concerned with making the mode of operation as obvious to the child as possible, for example only using RoboTed's speech synthesis when in autonomous operation.

\paragraph{Over trusting by parents} 
The risk here is that parents become over reliant on RoboTed’s child minder function \cite{feil2011socially}. The risk and its consequences are so great as to suggest the child minder function should be removed altogether.

\paragraph{The Uncanny Valley} 
The Uncanny Valley can lead to a fearful reaction when a robot is close to but not 100\% lifelike \cite{moor12}. The risk of this is probably low with RoboTed, both because RoboTed is not human-like at all, and children are already familiar with teddy bears. However, the risk should be explored by engaging children in early trials of RoboTed, and if the uncanny value reaction is demonstrated it might be mitigated by, for instance, equipping the robot with a ‘cartoon’ voice.

\subsubsection{Privacy and Security Risks}

We next consider privacy and transparency risks, summarised in Table \ref{tab:3}.

\begin{table}
\tbl{Privacy and Security Risk Analysis for RoboTed}
{\begin{tabular}{p{2cm} p{4cm} p{1cm} p{4cm}}\toprule
Hazard & Risk & Level$^{\text a}$ & Mitigation  \\
\colrule
Weak security & Malicious hackers gain access to RoboTed’s sensors \& control functions & H & Implement strong encryption together with best practice password protection\\
Privacy & Personal data, including images and voice recordings of child are stolen & M & Put in place auditable measures to ensure personal data is deleted immediately\\
Lack of transparency & Lack of data logs makes it hard or impossible to investigate accidents & H & Build a secure local data logger into RoboTed\\
\botrule
\end{tabular}
}
\begin{tabnote}
$^{\text a}$ (H)igh, (M)edium or (L)ow\\
\end{tabnote}
\label{tab:3}
\end{table}

\paragraph{Weak security}
Weak security could lead to malicious hackers gaining access to RoboTed’s sensors \& control functions. This could be very frightening for a child and her parents. To reduce the risk we need to implement strong encryption of the communications between RoboTed and the cloud, alongside best practice password protection to make it very hard for hackers to guess the password.

\paragraph{Privacy} 
Here the risk is that personal data, including images and voice recordings of children (and the house they live in) are stolen. One way of reducing this risk would be to ensure that personal data sent to the cloud is deleted immediately after it has been used.

\paragraph{Lack of transparency} 
The risk is that if there were an accident in which RoboTed harmed a child –- noting that it could be either physical or psychological harm –- it would be very difficult to investigate what happened to cause the accident unless the robot keeps a data log of its actions and responses. This is a serious risk and to mitigate the risk a secure data logger –- \textit{ethical black box} –- needs to be built into RoboTed \cite{winf17}. The data would only be stored locally, and only the most recent few hours of data would need to be saved. 

\subsubsection{Environmental Risks}

Finally we consider the environmental risks summarised in Table \ref{tab:4}.

\begin{table}
\tbl{Environmental Risk Analysis for RoboTed}
{\begin{tabular}{p{2cm} p{4cm} p{1cm} p{4cm}}\toprule
Hazard & Risk & Level$^{\text a}$ & Mitigation  \\
\colrule
Unsustainability (of materials) & Robot uses unsustainable or high carbon cost materials & M & Use materials (e.g. RoboTed’s fur) from sustainable sources, avoiding plastics.\\
Unrepairability & Robot’s lifetime is limited because faults cannot be repaired or parts replaced & M & Design for ease of repair with replaceable parts -- especially battery\\
Unrecyclability & End of life robots are dumped in land fill & M & Design for ease of recycling parts and materials\\
\botrule
\end{tabular}
}
\begin{tabnote}
$^{\text a}$ (H)igh, (M)edium or (L)ow\\
\end{tabnote}
\label{tab:4}
\end{table}

\paragraph{Unsustainability of materials} 
Here the risk is that the robot uses unsustainable or high carbon materials. To mitigate this risk we could use materials (e.g. RoboTed’s fur) from sustainable sources. We could also avoid plastics by, for instance, using wood for RoboTed’s skeleton.

\paragraph{Unrepairability}
This leads to the risk that the robot’s lifetime is limited because faults cannot be repaired or parts replaced. This risk can be minimized by designing RoboTed for ease of repair, using replaceable parts as much as possible –- especially the battery. Additionally RoboTed’s manufacturers should provide a repair manual so that local workshops can fix most faults.

\paragraph{Unrecyclability}
All products will eventually come to the end of their useful life, and if they cannot be repaired or recycled we risk them being dumped in landfill. To mitigate this risk RoboTed should be designed to make it easy to recycle parts. Ideally after these parts have been recovered for recycling the remaining materials are biodegradable.

\section{Discussion and conclusions}

Through a fictional case study this paper has demonstrated the value of ethical risk assesment. Our RoboTed ERA has shown that attention to ethical risks can 
\begin{itemize}
    \item suggest new functions, such as “RoboTed needs to sleep now”,
    \item draw attention to how designs can be modified to mitigate some risks, 
    \item highlight the need for user engagement, and
    \item reject some product functionality as too risky.
\end{itemize}

ERA is, however, not guaranteed to expose all ethical risks. It is a subjective process which will only be successful if the risk assessment team are prepared to think both critically and creatively about the question: what could go wrong? As Vallor \textit{et al} write\cite{vall2018}, design teams must develop the ``habit of exercising the skill of \textit{moral imagination} to see how an ethical failure of the project might easily happen, and to understand the preventable causes so that they can be mitigated or avoided". 

The ethical hazards and risks set out in BS8611 are an excellent starting point, but the standard does not provide an exhaustive taxonomy of ethical hazards, encompassing all domains of robotics. Our RoboTed case study has identified several additional ethical hazards, some of which are specific to social robots, including the Uncanny Valley, weak security, lack of transparency (for instance the lack of data logs needed to investigate accidents), unrepairability and unrecyclability.  

Evaluating and quantifying psychological risks is especially difficult given that there are no agreed measures for hazards such as over trusting or the Uncanny Valley. Assessment is further complicated by the likelihood that cultural differences may lead to lower risks of psychological harm to some individuals than others. For these reasons design teams cannot rely on their own judgement and instead should engage with potential users representative of the full range of age, gender and ethnic diversity, and seek the guidance of psychologists and/or social scientists both to ask the user group the right questions and interpret their responses. Given also that ERA is not a one-time process but one that should iterate through a product life-cycle, good practice suggests that in-house assessments undertaken early in the design process would be shared with user groups during later iterations as the product undergoes user trials.

In summary ethical risk assessment is a powerful and essential addition to the responsible roboticist's toolkit. ERA can also be thought of as the opposite face of robot accident investigation\cite{winf2020}, seeking –- at design time –- to prevent risks becoming accidents.

\section*{Acknowledgments}
The work of this paper has been conducted within EPSRC project Robo-TIPS, grant reference EP/S005099/1 \textit{RoboTIPS: Developing Responsible Robots for the Digital Economy} The authors are also grateful for the comments of the anonymous reviewers.

\bibliographystyle{ieeetr}
\bibliography{ICRES-sample}

\end{document}